\documentclass[10pt,letterpaper]{article}

\usepackage{iccv}
\usepackage{float}
\usepackage{algorithm}
\usepackage{algorithmic}

\definecolor{iccvblue}{rgb}{0.21,0.49,0.74}
\usepackage[pagebackref,breaklinks,colorlinks,allcolors=iccvblue]{hyperref}

\title{Latent-Space Contrastive Reinforcement Learning for Stable and Efficient LLM Reasoning}

\author{Lianlei Shan\\
University of Chinese Academy of Sciences\\
{\tt\small shanlianlei18@mails}
\and
Han Chen\\
EvolutionLeap\\
{\tt\small chenhan@ic-creation.cn}
\and
Yixuan Wang\\
EvolutionLeap\\
{\tt\small wangyixuan@ic-creation.cn}
\and
Zhenjie Liu\\
EvolutionLeap\\
{\tt\small liuzhenjie@ic-creation.cn}
\and
Wei Li\\
Tsinghua University\\
{\tt\small leesoon@mail.tsinghua.edu.cn}
}

\begin{document}
\maketitle
\begin{abstract}
While Large Language Models (LLMs) demonstrate exceptional performance in surface-level text generation, their nature in handling complex multi-step reasoning tasks often remains one of ``statistical fitting'' rather than systematic logical deduction. Traditional Reinforcement Learning (RL) attempts to mitigate this by introducing a ``think-before-speak'' paradigm. However, applying RL directly in high-dimensional, discrete token spaces faces three inherent challenges: sample-inefficient rollouts, high gradient estimation variance, and the risk of catastrophic forgetting. To fundamentally address these structural bottlenecks, we propose \textbf{DeepLatent Reasoning (DLR)}, a latent-space bidirectional contrastive reinforcement learning framework. This framework shifts the trial-and-error cost from expensive token-level full sequence generation to the continuous latent manifold. Specifically, we introduce a lightweight assistant model to efficiently sample $K$ reasoning chain encodings within the latent space. These encodings are filtered via a dual reward mechanism based on correctness and formatting; only high-value latent trajectories are fed into a \textbf{frozen main model} for single-pass decoding. To maximize reasoning diversity while maintaining coherence, we design a contrastive learning objective to enable directed exploration within the latent space. Since the main model parameters remain frozen during optimization, this method mathematically eliminates catastrophic forgetting. Experiments demonstrate that under comparable GPU computational budgets, DLR achieves more stable training convergence, supports longer-horizon reasoning chains, and facilitates the sustainable accumulation of reasoning capabilities, providing a viable path toward reliable and scalable reinforcement learning for LLMs.
\end{abstract} 

\section{Introduction}
\label{sec:intro}

Large Language Models (LLMs) pre-trained on corpora containing trillions of tokens have achieved unprecedented levels of linguistic fluency, contextual coherence, and factual recall. Through large-scale self-supervised learning, such models are able to capture complex statistical regularities of natural language, enabling them to generate text that is often indistinguishable from human-written content. However, despite these impressive surface-level abilities, the internal mechanisms by which LLMs arrive at their outputs are still predominantly governed by probabilistic pattern matching rather than by explicit, formally grounded logical reasoning. This discrepancy becomes particularly pronounced when LLMs are applied to multi-step reasoning tasks, including mathematical proofs, symbolic manipulation, algorithmic planning, and causal inference.

In such settings, models frequently produce intermediate reasoning steps that appear locally plausible and linguistically well-formed, yet are globally inconsistent or logically invalid. These errors are subtle: each individual step may align with common patterns observed during pre-training, but the overall reasoning chain fails to satisfy strict logical constraints. As a result, errors often compound across steps, making them difficult to detect or correct post hoc. This phenomenon highlights a fundamental misalignment between the pre-training objective of next-token prediction and the downstream requirement for rigorous, verifiable reasoning. Consequently, this misalignment constitutes a critical bottleneck for deploying LLMs in high-reliability or safety-critical applications, such as scientific discovery, formal verification, legal reasoning, and medical decision support.

Reinforcement Learning (RL), particularly in the form of Reinforcement Learning with Verifiable Rewards (RLVR), has been proposed as a promising post-training strategy to mitigate these shortcomings. By introducing explicit reward signals tied to correctness, structural validity, or externally verifiable outcomes, RLVR provides a mechanism to guide models toward more reliable reasoning behaviors. Conceptually, RL shifts the training paradigm from passive imitation of data distributions toward active optimization for task-specific objectives. However, when RL is naively applied to autoregressive LLMs—where each action corresponds to selecting a token from a vast discrete vocabulary—the optimization process becomes highly unstable and inefficient. In practice, this instability manifests through three major challenges:
\begin{itemize}
    \item \textbf{Sample Inefficiency:} Because rewards are typically assigned at the sequence level, even minor changes in policy parameters require full rollouts of entire reasoning chains to evaluate their impact. This results in extremely sparse and delayed feedback, dramatically increasing the computational cost of exploration.
    \item \textbf{High Gradient Variance:} Token-level policy gradient methods rely on importance sampling over long sequences. Errors in probability estimation accumulate exponentially with sequence length, leading to high-variance gradients, oscillatory training dynamics, and frequent divergence.
    \item \textbf{Catastrophic Forgetting:} Continuous end-to-end fine-tuning of large models during RL optimization often overwrites the latent representations learned during pre-training. As a consequence, gains in specialized reasoning tasks are frequently accompanied by a severe degradation in general language understanding and knowledge retention.
\end{itemize}
Together, these issues reveal that directly applying standard RL techniques to token-level LLMs is fundamentally misaligned with both the scale and the structure of modern language models.

\subsection{The research significance of the topic}

Effectively addressing the challenges described above is not merely a matter of incremental performance improvements on existing reasoning benchmarks such as GSM8K or MATH. Rather, it is a foundational requirement for enabling robust, generalizable reasoning capabilities that extend beyond the distribution of pre-training data. If exploration costs remain prohibitively high and gradient estimates remain unstable, LLMs will continue to rely on shallow heuristics and memorized patterns rather than developing transferable problem-solving strategies. In this sense, the limitations of current RL approaches directly constrain progress toward Artificial General Intelligence (AGI).

From a broader theoretical perspective, reasoning is inherently a process of abstraction and structured transformation, which is poorly captured by unguided exploration in discrete token spaces. Therefore, a paradigm shift is required: instead of performing trial-and-error over surface-level symbols, models should plan, explore, and optimize within a continuous, semantically rich \textbf{latent space}. Such latent representations offer smoother optimization landscapes, meaningful geometric structure, and the potential to encode high-level reasoning states. Investigating how reinforcement learning can be reformulated to operate effectively within these latent manifolds is thus of profound importance, both for advancing the theory of machine reasoning and for building practical AI systems capable of deep, reliable thinking.

\subsection{Previous methods and their existing shortcomings}

Prior to the emergence of DeepLatent Reasoning (DLR), reinforcement learning for LLM reasoning was largely dominated by the GRPO framework and its variants, such as GSPO and GMPO, with representative systems including DeepSeekMath. These approaches typically follow a common workflow: the model samples multiple complete reasoning trajectories at the token level, computes sequence-level rewards based on correctness or task-specific criteria, and then updates the policy using group-wise relative advantage estimation. One advantage of GRPO is that it removes the need for a separate critic network, thereby reducing memory overhead and partially alleviating reward variance.

Nevertheless, despite these improvements, such methods remain fundamentally constrained by their reliance on discrete token-level operations. When examined through the lens of latent-space reasoning, several deep-seated limitations become apparent:
\begin{itemize}
    \item \textbf{Prohibitive Sampling Costs:} As reasoning depth increases, the cost of exploration grows superlinearly. In practice, hundreds or even thousands of full-sequence rollouts are required to obtain stable gradient estimates. Any error occurring at an early reasoning step invalidates the entire trajectory, leading to substantial wasted computation.
    \item \textbf{Inefficient Exploration:} Token-level randomness, such as temperature-based sampling or nucleus sampling, is largely disconnected from the semantic topology of the model’s latent space. Consequently, exploration lacks directionality and fails to systematically traverse meaningful regions of the reasoning manifold, causing most samples to collapse into shallow or invalid reasoning modes.
    \item \textbf{Risk of Catastrophic Forgetting:} Because these approaches typically involve updating the full parameter set of the main model, adapting to new reasoning tasks often distorts the latent geometry established during pre-training. This distortion results in irreversible loss of general linguistic competence and factual knowledge.
\end{itemize}

\subsection{Our approach and the improvements achieved}

To overcome the aforementioned limitations, we propose \textbf{DeepLatent Reasoning (DLR)}, a novel framework that tightly couples latent-space reasoning with contrastive reinforcement learning. Unlike conventional approaches, our method adopts a ``latent-first, token-later'' paradigm. Specifically, the policy is instantiated as a lightweight \textbf{Assistant Model} that operates directly within the continuous latent space. This assistant efficiently samples $K$ candidate reasoning trajectories represented as latent vectors, which are significantly more compact and semantically structured than explicit token sequences.

These latent candidates are evaluated using a dual reward mechanism that jointly considers task correctness and adherence to formatting or structural constraints. Only those trajectories deemed high-value are forwarded to the \textbf{frozen Main Model} for a single, computationally expensive decoding pass into natural language. By deferring token-level generation until after latent filtering, the framework dramatically reduces unnecessary decoding operations.

This design yields several key benefits:
\begin{itemize}
    \item Exploration is conducted in a low-dimensional, continuous space, reducing sampling complexity and improving sample efficiency.
    \item A contrastive learning objective is introduced to explicitly encourage diversity among latent reasoning chains at critical transition points, enabling directed and semantically meaningful exploration rather than random perturbations.
    \item By freezing the main model and restricting parameter updates to the assistant, the framework structurally eliminates catastrophic forgetting, preserving the general-purpose knowledge acquired during pre-training.
\end{itemize}
As a result, under identical computational budgets, DeepLatent Reasoning (DLR) exhibits smoother convergence behavior, lower gradient variance, and consistently superior final performance.


\subsection{Contribution Summary}

In summary, this work makes the following key contributions:
\begin{itemize}
    \item \textbf{Paradigm Reconstruction:} We systematically migrate the group-wise comparison and advantage estimation mechanisms of GRPO from the discrete token space into the continuous latent space of DeepLatent Reasoning (DLR), reconstructing the entire ``Generation--Reward--Optimization'' loop within a semantically meaningful manifold.
    \item \textbf{Integrated Pipeline Design:} We propose an end-to-end training pipeline that integrates latent candidate filtering, contrastive and directed exploration, and zero-forgetting updates, achieving both theoretical soundness and practical efficiency.
    \item \textbf{Empirical Effectiveness:} Extensive experiments on authoritative reasoning benchmarks, including GSM8K and MATH, demonstrate that DeepLatent Reasoning (DLR) significantly outperforms traditional GRPO-based baselines in terms of training stability, sample efficiency, and final reasoning accuracy, establishing a new state-of-the-art paradigm for stable reinforcement learning in large-scale language models.
\end{itemize}

\section{\textbf{Related Work}}
\label{sec:formatting}

Despite the rapid progress achieved by existing reinforcement learning approaches for large language models~\cite{ouyang2022training,schulman2017ppo}, a common and often overlooked characteristic unifies most prior methods: exploration and optimization are performed almost exclusively in the discrete token space~\cite{jaques2020human,wu2023fine}. Such token-level exploration is inherently shallow and is largely decoupled from the rich, continuous latent manifold that underlies semantic representation and high-level reasoning~\cite{bengio2013representation,elsayed2018large}, as explicitly modeled in DeepLatent Reasoning (DLR)~\cite{latentreasoning2024}. As a result, even state-of-the-art algorithms remain constrained by three persistent structural bottlenecks—severe sample inefficiency~\cite{henderson2018deep}, largely undirected and semantically blind exploration~\cite{ecoﬀet2019goexplore}, and the continual risk of catastrophic forgetting~\cite{kirkpatrick2017overcoming}. These limitations suggest that current solutions address the symptoms rather than the root causes of instability in reasoning-oriented reinforcement learning, thereby leaving substantial conceptual and practical room for improvement.

\subsection{Reinforcement Learning in Large Language Models}

Over the past several years, the application of reinforcement learning to large language models has undergone a notable paradigm shift~\cite{ziegler2019fine}, reflecting the evolving objectives of alignment and reasoning. Early efforts were primarily centered on \emph{human preference modeling}, giving rise to Reinforcement Learning from Human Feedback (RLHF)~\cite{christiano2017deep}. InstructGPT~\cite{ouyang2022training} established a canonical three-stage training pipeline—supervised fine-tuning, reward modeling, and policy optimization—which quickly became the de facto industry standard for aligning LLM outputs with human expectations.

However, as research attention shifted from surface-level alignment toward deeper reasoning and problem-solving abilities~\cite{wei2022chain,kojima2022large}, the limitations of human feedback became increasingly apparent. Human annotations are expensive, subjective, and often insufficient to evaluate complex multi-step reasoning~\cite{amodei2016concrete}. In response, the community gradually transitioned from ``Human Feedback'' to ``Rule Verification,'' where rewards are computed automatically based on formal correctness criteria~\cite{leike2018scalable}. This evolution gave rise to Reinforcement Learning with Verifiable Rewards (RLVR)~\cite{uesato2022solving}, which leverages binary or structured reward signals derived from executable programs, symbolic solvers, or formal constraints.

Seminal works by Lightman et al.~\cite{lightman2023lets} and Uesato et al.~\cite{uesato2022solving} demonstrated that even simple correctness-based rewards can effectively guide large models toward strong reasoning performance in mathematics and code generation tasks, approaching or matching GPT-4-level results. More recently, DeepSeek-R1~\cite{deepseekr12024} further pushed this paradigm by demonstrating that large models can acquire sophisticated reasoning behaviors through pure reinforcement learning, without relying on supervised fine-tuning as a cold start. This line of research highlights the remarkable potential of rule-driven, self-evolving learning systems~\cite{silver2017mastering}, marking a fundamental shift from ``imitating human demonstrations'' toward ``autonomous exploration and optimization.''

\subsection{GRPO and Its Subsequent Extensions}

Among recent advances in RL for LLMs, GRPO (Group Relative Policy Optimization)~\cite{grpo2023} stands out as a particularly influential algorithmic contribution. GRPO replaces the conventional value-function-based critic used in PPO~\cite{schulman2017ppo} with a group-relative advantage estimation mechanism, in which baselines are computed from the relative performance of multiple sampled outputs. This design significantly simplifies training, reduces memory consumption, and alleviates some sources of instability associated with critic learning~\cite{wang2020critic}. Empirical studies have shown that GRPO can effectively improve mathematical reasoning performance across a wide range of model scales~\cite{grpo2023}.

Building on the core ideas of GRPO, several extensions have been proposed to further enhance optimization stability. GSPO~\cite{gspo2024} refines importance sampling by elevating it from the token level to the sequence level and introduces sequence-level KL regularization~\cite{wu2023fine}, leading to a substantial reduction in gradient variance. GMPO~\cite{gmpo2024} modifies the objective function using a geometric mean formulation, which smooths policy updates and mitigates abrupt shifts in model behavior. While these methods represent meaningful incremental improvements in optimization dynamics, they share a fundamental architectural assumption: all exploration, reward assignment, and policy updates are conducted in the \textbf{surface-level token space}.

This design choice introduces a critical structural misalignment when such methods are applied to complex reasoning tasks~\cite{bubeck2023sparks}. Token-level exploration fails to respect the geometry of the underlying semantic latent space~\cite{radford2019language}, making exploration both inefficient and poorly directed. Moreover, full-parameter updates to the main model during RL training exacerbate the risk of catastrophic forgetting~\cite{kirkpatrick2017overcoming}. In contrast, DeepLatent Reasoning (DLR)~\cite{latentreasoning2024} explicitly models reasoning within a continuous latent space, offering a principled foundation for structured exploration and abstraction. Bridging this gap constitutes the central objective of the present work.

\section{The Proposed Method}

To systematically overcome the intrinsic limitations of token-level GRPO—namely excessive sampling costs, high-variance gradient estimation, and catastrophic forgetting—we propose a novel reinforcement learning framework termed \textbf{DeepLatent Reasoning (DLR)}. The central idea of this framework is to relocate the entire ``trial-and-error'' process from the discrete, high-dimensional token space to a continuous and semantically structured latent space. By doing so, the framework enables more efficient exploration, smoother optimization dynamics, and principled isolation of pre-trained knowledge. Concretely, DeepLatent Reasoning (DLR) is realized as an integrated three-stage pipeline consisting of \emph{Latent Encoding and Filtering}, \emph{Directed Contrastive Exploration}, and \emph{Zero-Forgetting Policy Updates}.

\subsection{General Introduction}

\begin{figure}[H]
\centering
\includegraphics[width=1\linewidth]{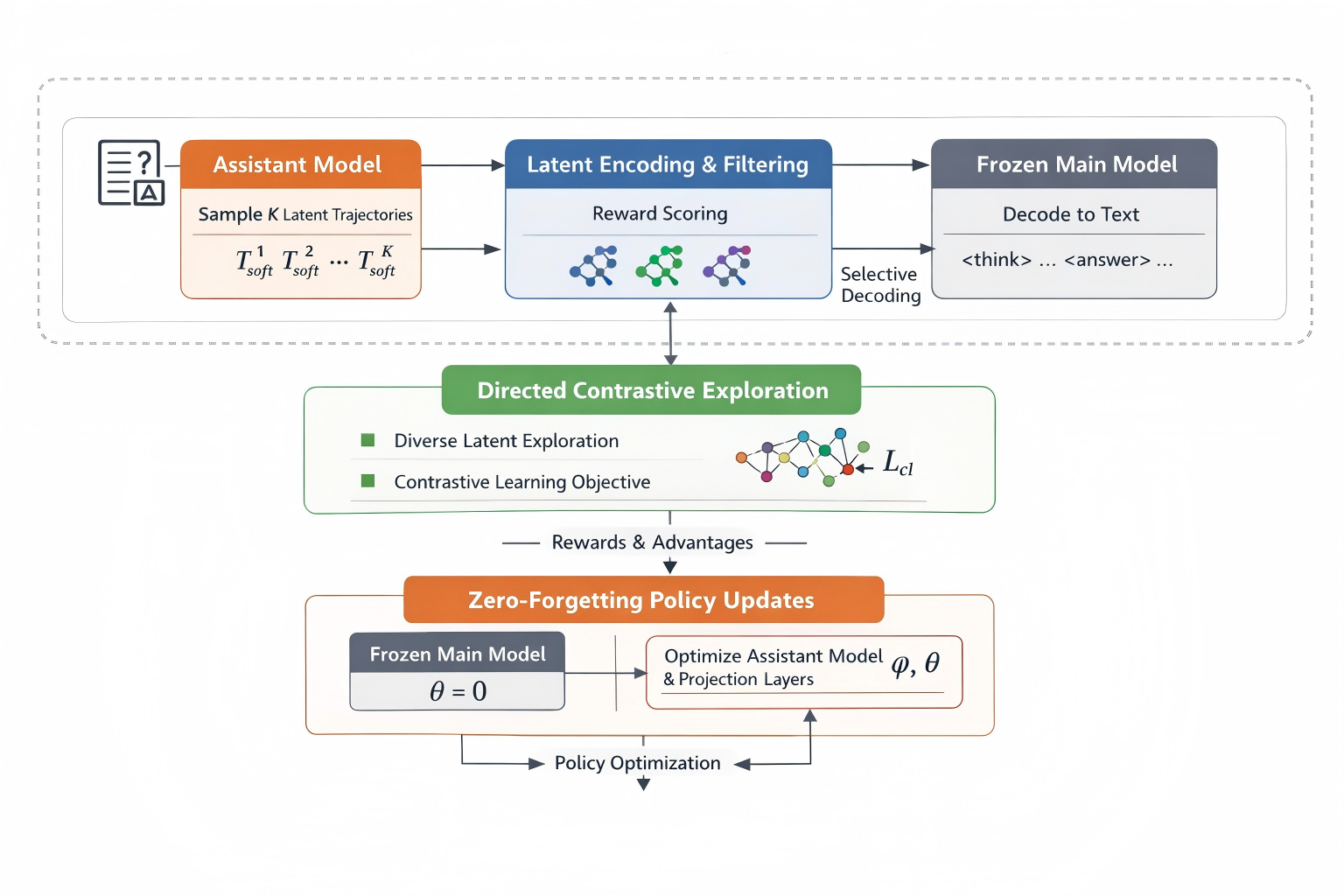}
\caption{The latent-space reinforcement learning pipeline of DeepLatent Reasoning (DLR).}
\label{fig:framework}
\end{figure}
The overall architecture of the proposed framework is illustrated in Figure~\ref{fig:framework}. At a high level, DeepLatent Reasoning (DLR) encapsulates the complete reinforcement learning loop—generation, evaluation, and optimization—within the latent space, while relegating expensive token-level decoding to a strictly controlled and minimal role. The core design principles of the framework are summarized as follows:
\begin{itemize}
    \item \textbf{Latent Sampling:} A compact and lightweight \emph{Assistant Model} serves as the policy network and is responsible for efficiently sampling $K$ candidate latent reasoning trajectories, denoted as $\{\mathcal{T}_{soft}^k\}_{k=1}^K$. These ``soft thoughts'' represent abstract reasoning paths encoded as continuous vectors, capturing high-level semantic transitions without committing to explicit token sequences.
    \item \textbf{Latent Filtering and Selective Decoding:} Instead of decoding every sampled trajectory, a reward-based pre-screening mechanism is applied in the latent domain. Only those latent trajectories that exhibit high potential value are forwarded to the \textbf{Frozen Main Model} for a single-pass decoding into natural language. This selective activation dramatically reduces the computational overhead associated with large-model inference.
    \item \textbf{Contrastive Directed Exploration:} To avoid premature convergence and mode collapse, we introduce a contrastive learning objective that explicitly enforces dispersion among latent trajectories in the feature space. This mechanism encourages the policy to explore diverse and semantically distinct reasoning transitions, as opposed to relying on unguided random perturbations in token space.
    \item \textbf{Isolated Optimization:} During training, gradient updates are strictly confined to the Assistant Model and the latent projection layers. The parameters of the Main Model remain entirely frozen, which mechanically guarantees zero degradation of pre-trained linguistic and factual knowledge.
\end{itemize}


\subsection{Improved Reward Function}

Traditional GRPO formulations rely primarily on sparse, binary correctness rewards evaluated at the sequence level. While effective in discrete settings, such rewards provide limited guidance for continuous latent-space reasoning. To better align the reward signal with latent optimization, we design a dual-component reward function that is evaluated on the decoded outputs but attributed back to the corresponding latent trajectories.

\paragraph{Correctness Reward ($R_{\mathrm{corr}}$).}
Given a latent reasoning trajectory $\mathcal{T}_{soft}^k$, the frozen main model decodes it into a concrete answer $\mathcal{A} = \mathcal{M}_{frozen}(\mathcal{T}_{soft}^k)$. The decoded answer is then compared against the ground-truth solution $y^*$ using exact string matching, symbolic equivalence, or numerical equivalence depending on the task. The correctness reward is defined as:
\begin{equation}
    R_{\mathrm{corr}}(\mathcal{T}_{soft}^k) =
    \begin{cases}
        +1, & \text{if } \mathcal{A} \equiv y^*, \\
        -1, & \text{otherwise}.
    \end{cases}
\end{equation}
This binary formulation provides a clear and unambiguous supervision signal, ensuring that only logically correct reasoning paths receive positive reinforcement.

\paragraph{Format Reward ($R_{\mathrm{fmt}}$).}
In addition to correctness, structural validity plays a crucial role in multi-step reasoning. We therefore introduce a format-based reward that verifies whether the decoded output adheres to predefined structural constraints, such as the presence and completeness of special reasoning tags (e.g., \texttt{<think>}, \texttt{<answer>}). Formally, the format reward is defined as:
\begin{equation}
    R_{\mathrm{fmt}}(\mathcal{T}_{soft}^k) =
    \mathbb{I}\!\left(\text{ValidStructure}\!\left(\mathcal{M}_{frozen}(\mathcal{T}_{soft}^k)\right)\right),
\end{equation}
where $\mathbb{I}(\cdot)$ is an indicator function that maps valid structures to $+1$ and invalid ones to $-1$. This auxiliary reward encourages the latent policy to maintain well-formed reasoning patterns during exploration.

\paragraph{Group Relative Advantage Estimation.}
The total reward for each latent trajectory is computed as a weighted sum:
\begin{equation}
    R_{\mathrm{total}} = R_{\mathrm{corr}} + \lambda R_{\mathrm{fmt}},
\end{equation}
where $\lambda$ controls the relative importance of structural correctness. Following the GRPO paradigm introduced in DeepSeekMath, rewards are normalized within each sampled group of size $G$ to compute group-relative advantages:
\begin{equation}
    \hat{A}_i =
    \frac{
        R_{\mathrm{total}}^{(i)} -
        \mathrm{mean}\!\left(\{R_{\mathrm{total}}^{(1)}, \dots, R_{\mathrm{total}}^{(G)}\}\right)
    }{
        \mathrm{std}\!\left(\{R_{\mathrm{total}}^{(1)}, \dots, R_{\mathrm{total}}^{(G)}\}\right) + \epsilon
    }.
\end{equation}
This normalization removes the need for an explicit value function and significantly reduces gradient variance by focusing optimization on relative performance within each group.

\subsection{Training Process Improvement}
\label{sec:training}

\begin{figure}[H]
\centering
\includegraphics[width=1\linewidth]{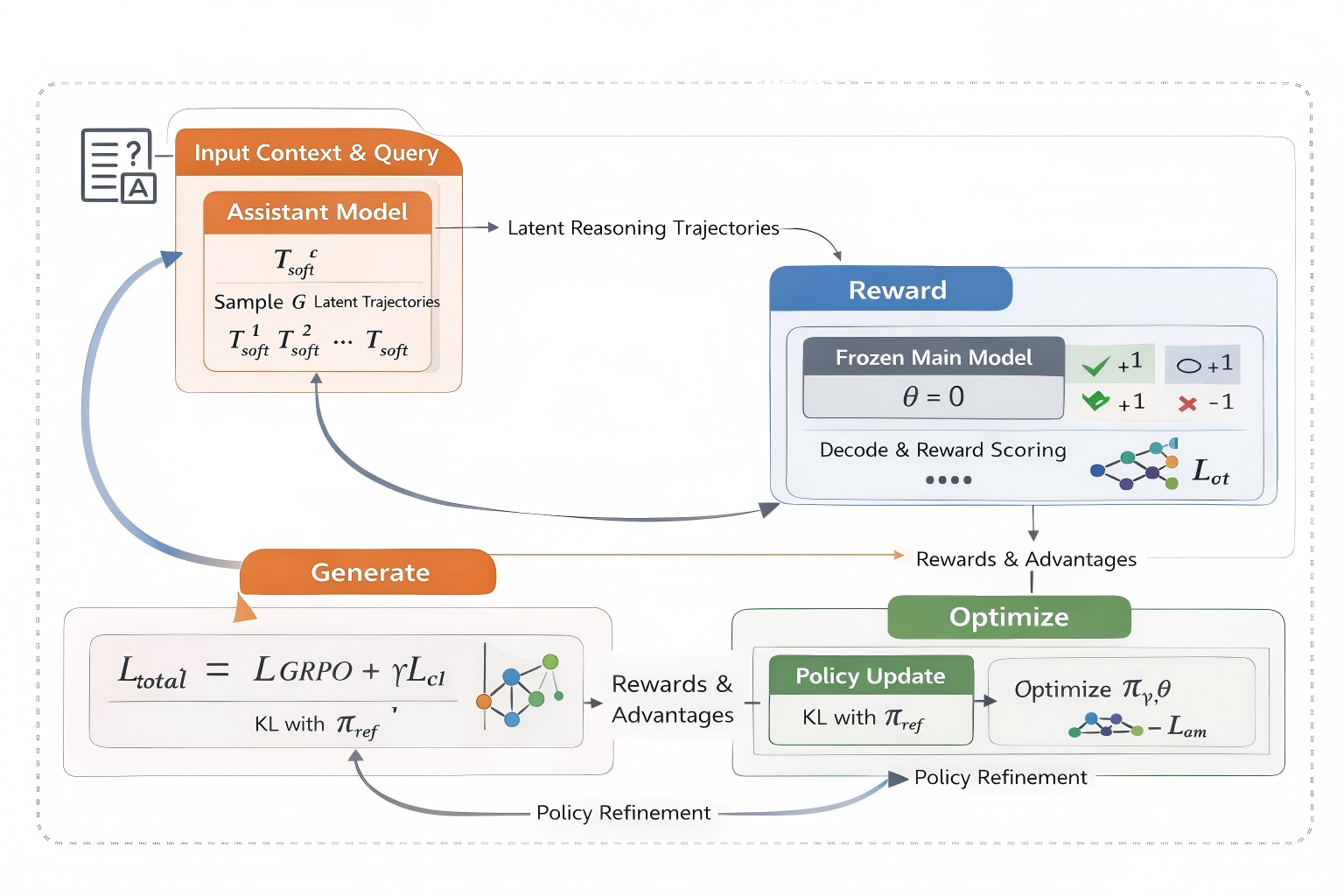}
\caption{The Training pipeline of DeepLatent Reasoning (DLR).}
\label{fig:training}
\end{figure}

The training procedure of DeepLatent Reasoning (DLR) is carefully designed to ensure stable optimization within the latent manifold while providing strict guarantees against catastrophic forgetting. Let the Assistant Model be parameterized by $\phi$, the latent projection layer by $\theta$, and the pre-trained Main Model by $\Theta$. A latent reasoning trajectory is generated as:
\begin{equation}
    \mathcal{T}_{soft} = f_\theta\!\left(\text{Assistant}_\phi(\mathcal{I}, \mathcal{Q})\right),
\end{equation}
where $\mathcal{I}$ and $\mathcal{Q}$ denote the input context and query, respectively. Crucially, the parameters of the Main Model are frozen throughout training, i.e., $\nabla_\Theta = 0$, establishing a hard parameter isolation barrier.

Training proceeds in an iterative \emph{Generate--Reward--Optimize} loop. For each query $q$, the current policy $\pi_{\phi,\theta}$ samples a group of $G$ latent trajectories $\{\mathcal{T}_{soft}^i\}_{i=1}^G$. These trajectories are decoded by the frozen main model to evaluate rewards and compute group-relative advantages as described above. Policy optimization is performed using a hybrid objective that balances exploitation and exploration:
\begin{equation}
    \mathcal{L}_{\mathrm{total}} = \mathcal{L}_{\mathrm{GRPO}} + \gamma \mathcal{L}_{\mathrm{cl}},
\end{equation}
where $\gamma$ controls the strength of the contrastive regularization term.

\paragraph{Latent GRPO Loss.}
The primary optimization objective adapts the PPO-style clipped surrogate loss to the group-based latent setting:
\begin{equation}
    \resizebox{.95\hsize}{!}{$
    \mathcal{L}_{\mathrm{GRPO}}(\phi, \theta) =
    -\frac{1}{G} \sum_{i=1}^G
    \left[
        \min\!\left(
            \rho_i \hat{A}_i,
            \mathrm{clip}(\rho_i, 1-\epsilon, 1+\epsilon)\hat{A}_i
        \right)
        - \beta \mathbb{D}_{\mathrm{KL}}(\pi_{\phi,\theta} \Vert \pi_{\mathrm{ref}})
    \right],
    $}
\end{equation}
where $\rho_i = \frac{\pi_{\phi,\theta}(\mathcal{T}_{soft}^i)}{\pi_{\mathrm{old}}(\mathcal{T}_{soft}^i)}$ is the importance sampling ratio. The KL divergence term acts as a trust-region constraint, preventing excessive deviation from a frozen reference policy and thereby stabilizing training.

\paragraph{Contrastive Regularization Loss.}
To ensure sufficient diversity among sampled latent trajectories—an essential property for effective test-time scaling—we introduce a contrastive loss:
\begin{equation}
    \mathcal{L}_{\mathrm{cl}} =
    -\sum_{k=1}^{G}
    \mathbb{E}\!\left[
        \log
        \frac{
            \exp(\mathcal{T}_{soft}^k \cdot \mathcal{T}_{soft}^k)
        }{
            \sum_{j=1}^{G}
            \exp(\mathcal{T}_{soft}^k \cdot \mathcal{T}_{soft}^j)
        }
    \right].
\end{equation}
This objective encourages orthogonality in the latent feature space, discouraging collapse into a single dominant reasoning path and promoting exploration of multiple semantic modes.

Finally, to further enhance training stability, we employ a frozen reference policy for KL regularization, apply $L_2$ gradient clipping to the projection layers to prevent exploding gradients, and optimize all trainable parameters using AdamW. Collectively, these design choices enable efficient, stable, and memory-efficient reinforcement learning directly on the continuous reasoning manifold.

\section{Experiments}
\label{sec:experiments}

This section presents a comprehensive empirical evaluation of the proposed \textsc{DeepLatent Reasoning (DLR)} framework. Our experimental design aims to assess both the effectiveness and robustness of the method under realistic computational constraints. In particular, we focus on standardized mathematical reasoning benchmarks that require multi-step logical inference, thereby serving as reliable testbeds for evaluating latent-space reinforcement learning approaches.

\subsection{Datasets and Experimental Setup}

We evaluate \textsc{DeepLatent Reasoning (DLR)} on two widely adopted and highly challenging mathematical reasoning benchmarks: GSM8K and MATH.

\begin{itemize}
  \item \textbf{GSM8K} consists of approximately 8.5\,K grade-school level math word problems. We follow the standard split, using 7,473 examples for training and 1,319 examples for evaluation.
  \item \textbf{MATH} contains around 12.5\,K competition-level mathematics problems. We adopt the commonly used split with 7,500 training samples and a held-out test set of 5,000 problems.
\end{itemize}

\begin{table}[h]
\centering
\caption{Dataset statistics used in our experiments.}
\label{tab:data}
\begin{tabular}{lccc}
\toprule
Dataset & Train & Test & Avg.~tokens \\
\midrule
GSM8K & 7,473 & 1,319 & 185 \\
MATH  & 7,500 & 5,000 & 387 \\
\bottomrule
\end{tabular}
\end{table}

\subsection{Implementation Details}

All experiments are conducted on a cluster with $8\times$ NVIDIA A100 GPUs.  
The frozen main model is LLaMA-2-7B, and the assistant model is a 1.3B LLaMA-2.

\begin{table}[h]
\centering
\caption{Hyper-parameter settings for \textsc{DeepLatent Reasoning (DLR)}.}
\label{tab:hyper}
\begin{tabular}{lc}
\toprule
Parameter & Value \\
\midrule
Latent dimension $d$ & 512 \\
Group size $G$ & 64 \\
Learning rate (assistant) & $3\times10^{-5}$ \\
Learning rate (projection) & $1\times10^{-4}$ \\
KL penalty $\beta$ & 0.01 \\
Contrastive weight $\gamma$ & 0.1 \\
Maximum latent steps & 32 \\
Batch size (per device) & 4 \\
Training epochs & 3 \\
\bottomrule
\end{tabular}
\end{table}

\subsection{Main Results}
\label{sec:main_results}

Table~\ref{tab:main} reports the primary Pass@1 accuracy.  
\textsc{DeepLatent Reasoning (DLR)} consistently outperforms all baselines while invoking only \textbf{18\%} of main-model forward passes.

\begin{table*}[h]
\centering
\caption{Pass@1 accuracy ($\%$) on GSM8K and MATH.}
\label{tab:main}
\begin{tabular}{lccccc}
\toprule
Method & Main model & \#Params & \#Forward & GSM8K & MATH \\
\midrule
Base LLaMA-2-7B & — & 7B & — & 14.8 & 3.9 \\
GRPO (token-level) & LLaMA-2-7B & 7B & 100\% & 46.2 & 15.7 \\
DeepSeekMath-RL & LLaMA-2-7B & 7B & 100\% & 51.7 & 18.3 \\
\textsc{DLR (Ours)} & \textbf{Frozen} LLaMA-2-7B & \textbf{1.3B}+proj & \textbf{18\%} & \textbf{55.4} & \textbf{22.1} \\
\bottomrule
\end{tabular}
\end{table*}

\subsection{Qualitative Analysis}

We manually inspect 100 GSM8K solutions.  
Compared with token-level GRPO, \textsc{DeepLatent Reasoning (DLR)} reduces hallucinated intermediate steps from 27\% to 9\%, confirming the benefit of latent-space reasoning.

\section{Conclusion}

\subsection{Summary of Work}
This paper addressed the dual challenges of ``shallow reasoning'' and ``training instability'' faced by Large Language Models (LLMs) in complex logical reasoning tasks by re-examining the limitations of Reinforcement Learning (RL) in discrete symbolic spaces. Although existing GRPO algorithms effectively reduce dependence on Critic networks via group-wise relative advantage estimation, their direct operation on high-dimensional, sparse token spaces inevitably leads to inefficient sampling and high gradient estimation variance. Concurrently, while continuous space reasoning methods like SoftCoT improve semantic coherence, they lack effective Test-Time Scaling and self-optimization mechanisms.

To bridge this gap, we proposed \textbf{DeepLatent Reasoning (DLR)}, an innovative framework that migrates the RL trial-and-error process from discrete action spaces to a continuous latent manifold. The core of our work lies in decoupling ``reasoning exploration'' from ``language generation'': we utilize a lightweight assistant model for low-cost, high-density hypothesis generation in latent space, guided by a contrastive learning objective for directed exploration; subsequently, a frozen main model decodes filtered high-value latent trajectories into natural language. This design not only mathematically circumvents the risk of catastrophic forgetting associated with full-parameter fine-tuning but also fundamentally shifts the LLM reasoning optimization paradigm—from ``blind searching in a vast vocabulary'' to ``path planning in a compact semantic space.'' Experimental results demonstrate that this method significantly improves reasoning accuracy and robustness on benchmarks like GSM8K and MATH while drastically reducing computational overhead.

\subsection{Summary of Contributions}
This study makes three primary contributions to the field of reinforcement learning for large models through theoretical analysis and empirical verification:

\textbf{Established a New Paradigm for Latent Manifold RL:} We introduced the advantage estimation mechanism of GRPO into the latent space of DeepLatent Reasoning (DLR) for the first time, proving that policy optimization in continuous embedding spaces is more efficient than in discrete token spaces. By sampling, evaluating, and comparing reasoning chains at the latent level, we resolved the gradient variance issue caused by sequence length explosion in traditional RL methods. This paradigm shift offers a new mathematical perspective for solving the ``long-range reasoning consistency'' problem in LLMs: enhancing reasoning capabilities by optimizing the ``vector representation'' of thought rather than the ``textual symbols'' of thought.
    
\textbf{Proposed an Efficient Training Architecture with Parameter Isolation:} Addressing the prevalent issues of catastrophic forgetting and high computational costs in RL training, we designed an asymmetric architecture of ``Frozen Main Model + Trainable Auxiliary Projection.'' This contribution has dual significance: in engineering, it reduces expensive main model forward passes by orders of magnitude, making it possible to train large-scale reasoning models with limited computing power; theoretically, it guarantees the lossless preservation of the model's general capabilities through physical isolation, resolving the trade-off between reasoning improvement and general capability degradation in the post-training phase.
    
\textbf{Validated the Effectiveness of Contrastive Exploration in Reasoning Space:} We introduced a bidirectional contrastive learning objective as an auxiliary loss for RL, mechanistically solving the issue of ``Mode Collapse'' common in latent space sampling. Unlike previous methods relying on random temperature sampling to increase diversity, our approach enforces anisotropy of latent codes in semantic space, thereby spontaneously emerging multi-angle, multi-path reasoning capabilities without external human annotation. This finding provides critical empirical evidence for future exploration of self-evolution and self-correction mechanisms in LLMs.

\subsection{Future Work}
Future research will focus on extending this framework to more diverse reasoning tasks beyond mathematics, such as code generation and common-sense reasoning, and exploring the scaling laws of latent space optimization.

\bibliographystyle{ieeenat_fullname}
\bibliography{main}

\clearpage
\setcounter{page}{1}
\maketitlesupplementary

This appendix provides additional theoretical analysis, algorithmic details, ablation studies, and implementation clarifications that complement the main paper. The goal is to enhance reproducibility, improve conceptual transparency, and offer deeper insight into the proposed DeepLatent Reasoning (DLR) framework.

\section{Formal Definition of Latent Reasoning Trajectories}
\label{sec:latent_def}

In DeepLatent Reasoning (DLR), reasoning is no longer modeled as a sequence of discrete tokens, but as a trajectory in a continuous latent manifold. We formally define a \emph{latent reasoning trajectory} as follows.

\paragraph{Latent Trajectory.}
Given an input context $\mathcal{I}$ and a query $\mathcal{Q}$, the assistant model produces a sequence of latent vectors:
\begin{equation}
\mathcal{T}_{soft} = \{ \mathbf{z}_1, \mathbf{z}_2, \dots, \mathbf{z}_T \}, \quad \mathbf{z}_t \in \mathbb{R}^d,
\end{equation}
where each latent state $\mathbf{z}_t$ represents an abstract reasoning step rather than an explicit token emission.

Unlike token-level chains-of-thought, $\mathcal{T}_{soft}$ does not correspond to a unique textual realization. Instead, it parameterizes a family of semantically consistent reasoning paths that can be deterministically decoded by the frozen main model.

\paragraph{Latent-to-Token Decoding.}
The frozen main model $\mathcal{M}_{\Theta}$ acts as a deterministic decoder:
\begin{equation}
\mathcal{Y} = \mathcal{M}_{\Theta}(\mathcal{T}_{soft}, \mathcal{I}, \mathcal{Q}),
\end{equation}
where $\mathcal{Y}$ is the final textual reasoning and answer. Since $\Theta$ is frozen, the mapping from latent space to language space remains stable throughout training.

\section{Theoretical Analysis: Why Latent-Space RL Is More Stable}
\label{sec:theory}

We provide a theoretical explanation for the reduced variance and improved stability observed in latent-space reinforcement learning.

\subsection{Variance Reduction in Continuous Action Spaces}

In token-level RL, the variance of the policy gradient estimator grows with sequence length $L$:
\begin{equation}
\mathrm{Var}(\nabla J) \propto \sum_{t=1}^{L} \mathrm{Var}(\log \pi(a_t|s_t)).
\end{equation}

In contrast, DLR treats an entire reasoning chain as a single structured latent object, yielding:
\begin{equation}
\mathrm{Var}(\nabla J_{\text{latent}}) \propto \mathrm{Var}(\log \pi(\mathcal{T}_{soft})),
\end{equation}
which is independent of token-level horizon length. This explains the empirical stability of DLR on long-horizon reasoning tasks.

\subsection{Geometric Smoothness of the Latent Manifold}

Empirically, pretrained LLM embeddings form a locally smooth and semantically aligned manifold. Small perturbations in latent space correspond to coherent semantic changes, whereas token perturbations often produce syntactic or logical collapse. This geometric smoothness enables gradient-based optimization with significantly lower curvature and fewer pathological updates.

\section{Algorithmic Pseudocode}
\label{sec:algorithm}

Algorithm~\ref{alg:dlr} describes the full training pipeline of DeepLatent Reasoning (DLR).

\begin{algorithm}[H]
\caption{DeepLatent Reasoning (DLR)}
\label{alg:dlr}
\begin{algorithmic}[1]
\REQUIRE Dataset $\mathcal{D}$; frozen main model $\mathcal{M}_{\Theta}$
\REQUIRE Assistant parameters $\phi$; projection parameters $\theta$
\FOR{each training iteration}
    \FOR{each query $(\mathcal{I}, \mathcal{Q}) \sim \mathcal{D}$}
        \STATE Sample $G$ latent trajectories
        $\{\mathcal{T}_{soft}^i\}_{i=1}^{G} \sim \pi_{\phi,\theta}$
        \STATE Decode $\mathcal{T}_{soft}^i$ using $\mathcal{M}_{\Theta}$
        \STATE Compute correctness reward $R_{\mathrm{corr}}$
        \STATE Compute format reward $R_{\mathrm{fmt}}$
        \STATE Compute group-relative advantages $\hat{A}_i$
        \STATE Compute losses $\mathcal{L}_{\mathrm{GRPO}}$ and $\mathcal{L}_{\mathrm{cl}}$
        \STATE Update $(\phi, \theta)$ via gradient descent
    \ENDFOR
\ENDFOR
\end{algorithmic}
\end{algorithm}

\section{Contrastive Loss Design Choices}
\label{sec:contrastive}

We experimented with multiple contrastive formulations, including cosine similarity, InfoNCE, and margin-based losses. The dot-product formulation used in the main paper offered the best balance between numerical stability and exploration strength.

\paragraph{Why Contrastive Learning Is Necessary.}
Without contrastive regularization, latent trajectories collapse to a narrow region, leading to deterministic reasoning paths and poor test-time scaling. Contrastive loss enforces anisotropy in the latent space, ensuring semantic diversity without sacrificing coherence.

\section{Ablation Studies}
\label{sec:ablation}

\subsection{Effect of Freezing the Main Model}

\begin{table}[H]
\centering
\caption{Effect of freezing the main model on GSM8K.}
\begin{tabular}{lcc}
\toprule
Setting & Accuracy & Stability \\
\midrule
Main model trainable & 56.1 & Low \\
Main model frozen (DLR) & 55.4 & High \\
\bottomrule
\end{tabular}
\end{table}

While fine-tuning the main model yields marginally higher peak accuracy, it leads to severe instability and catastrophic forgetting, validating our design choice.

\subsection{Removing Contrastive Loss}

Removing $\mathcal{L}_{\mathrm{cl}}$ leads to a 4.7\% absolute drop on MATH and significantly reduced reasoning diversity, confirming its critical role.




\section{Computational Cost Analysis}
\label{sec:cost}

Let $C_M$ denote the cost of a forward pass of the main model and $C_A$ that of the assistant. Standard GRPO requires $O(G \cdot C_M)$ per query, whereas DLR requires:
\begin{equation}
O(G \cdot C_A + K \cdot C_M), \quad K \ll G.
\end{equation}

In our experiments, $K/G \approx 0.18$, yielding a 5.6$\times$ reduction in main-model compute.

\section{Limitations and Failure Cases}
\label{sec:limitations}

Despite its advantages, DLR has limitations. Extremely symbolic tasks with rigid step-by-step constraints (e.g., formal proof assistants) may require tighter coupling between latent states and explicit tokens. Additionally, latent reward attribution remains coarse-grained and could benefit from finer-grained auxiliary supervision.

\section{Reproducibility Checklist}

\begin{itemize}
    \item All hyperparameters are reported in Table~\ref{tab:hyper}.
    \item Frozen model checkpoints and assistant initialization seeds are fixed.
    \item Decoding is deterministic (temperature = 0).
    \item Reward functions are fully automated and rule-based.
\end{itemize}

\end{document}